\documentclass[a4paper,conference]{IEEEtran}

\usepackage{cite}
\usepackage{microtype}
\usepackage{graphicx}
\usepackage{subfigure}
\usepackage{booktabs} 
\usepackage[english]{babel}
\usepackage{blindtext}
\usepackage{balance}
\usepackage{threeparttable}

\def\sub#1{_{\rm #1}}

\def\eg{{\it e.g.}}

\def\ie{{\it i.e.}}

\usepackage{hyperref}

\begin{document}

\title{Adaptive Distillation for Decentralized Learning from Heterogeneous Clients}

\author{\IEEEauthorblockN{Jiaxin Ma}
\IEEEauthorblockA{OMRON SINIC X\\
Tokyo, Japan\\
Email: jiaxin.ma@sinicx.com}
\and
\IEEEauthorblockN{Ryo Yonetani}
\IEEEauthorblockA{OMRON SINIC X\\
Tokyo, Japan\\
Email: ryo.yonetani@sinicx.com}
\and
\IEEEauthorblockN{Zahid Iqbal}
\IEEEauthorblockA{University Sains Malaysia\\
Pulau Pinang, Malaysia\\
Email: zahid@student.usm.my}}

\maketitle

\begin{abstract}
This paper addresses the problem of decentralized learning to achieve a high-performance global model by asking a group of clients to share local models pre-trained with their own data resources. We are particularly interested in a specific case where both the client model architectures and data distributions are diverse, which makes it nontrivial to adopt conventional approaches such as Federated Learning and network co-distillation. To this end, we propose a new decentralized learning method called Decentralized Learning via Adaptive Distillation (DLAD). Given a collection of client models and a large number of unlabeled distillation samples, the proposed DLAD 1) aggregates the outputs of the client models while adaptively emphasizing those with higher confidence in given distillation samples and 2) trains the global model to imitate the aggregated outputs. Our extensive experimental evaluation on multiple public datasets (MNIST, \mbox{CIFAR-10}, and \mbox{CINIC-10}) demonstrates the effectiveness of the proposed method.
\end{abstract}

\section{Introduction}
Training high-performance deep neural networks typically requires a large-scale and diverse dataset. This requirement becomes challenging when we address supervised learning tasks on private data. As supervised tasks require every single training sample to be annotated with its ground truth label, an earlier work proposed leveraging crowdsourcing platforms to mitigate such high annotation costs~\cite{hsueh-etal-2009-data}. However, as the training data are assumed to be made public, this approach is not applicable for certain data that the owners wish to keep private, such as life-logging videos~\cite{Chowdhury2016}, biological data~\cite{Ching2018}, and medical data~\cite{Li2005}.

\begin{figure}[t]
  \begin{center}
    \includegraphics[width=\linewidth]{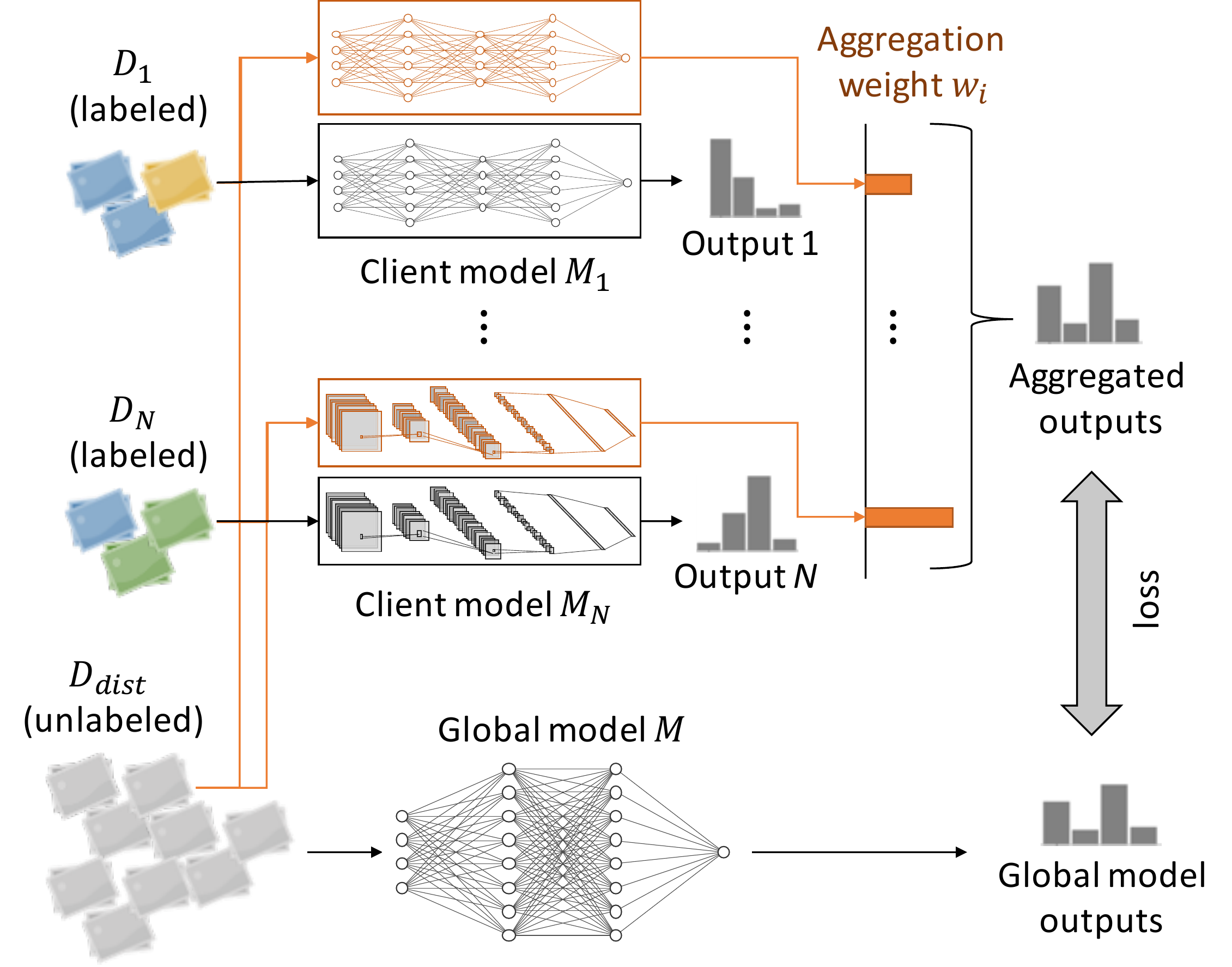}
    \caption{\textbf{Decentralized learning with DLAD.} Given a collection of client models with distinct architectures and trained with dissimilar sets of classes, we aim to achieve a global model that can recognize all the classes client data involve, without direct access to the data themselves.}
    \label{fig:overview}
  \end{center}
\end{figure}

To overcome this problem, a promising approach is decentralized learning, which asks a population of clients to train a local model with their private data resources and learns a target model (hereafter, global model) by aggregating the trained local models at a certain server. Such approaches make it possible to outsource the data collection and annotation processes while also allowing data to be kept private in the storage of original clients. One of the most popular frameworks is Federated Learning (FL)~\cite{Mcmahan2017,Bonawitz2019}, which iteratively conducts these local model training and global model aggregation steps. While FL is confirmed to work effectively for practical tasks that involve learning from large-scale private data~\cite{Hard2018,Wenqi2019}, we argue that it has several limitations.
\begin{enumerate}
\item FL requires each client to train a local model of the identical architecture. This requirement is practical for scenarios where clients are all equipped with the same hardware and software (\eg, smartphone devices of a similar spec with the latest OS). However, there are other scenarios where participating clients are heterogeneous~\cite{Bonawitz2019,nishio2019client} and allowed to train models with different architectures depending on their specs.
\item FL requires clients to regularly communicate with the server to exchange models. This makes it hard to utilize the approach in scenarios where clients are usually offline due to security reasons or poor network conditions and only limited communications are allowed upon request, such as learning from private data in factories or hospitals decentralized all over the world.
\end{enumerate}

To this end, we propose a new decentralized learning method that asks clients only once to submit their trained models. We leverage the idea of network distillation~\cite{hinton2015distilling} originally developed to transfer recognition abilities from one network to another by imitating network outputs. In the context of decentralized learning, we learn a global model to imitate outputs from client models. Doing so allows clients to train distinct models, while keeping their data locally, and submitting the model only once for training the global model. Nevertheless, adopting distillation methods to decentralized learning comes with two key technical challenges: a) a large-amount of annotated data are required for distilling client models, while such data are not assumed to be available on public in the decentralized learning scenarios; b) client data have been assumed to be identically distributed~\cite{ahn2019wireless,li2019fedmd}, whereas much work on decentralized learning is targeted at non-independent-and-identically-distributed (non-IID) data~\cite{Mcmahan2017}.

In order to address the aforementioned two challenges, we have developed \emph{Decentralized Learning via Adaptive Distillation (DLAD)}, which can accept data used in the distillation process to be \emph{unlabeled} and client models to be trained with data resources. As shown in Figure~\ref{fig:overview}, DLAD aggregates outputs from the client models to the distillation data adaptively such that 1) when client models have been trained with similar data to a given distillation sample, their outputs are regarded as `confident' ones and emphasized with higher weights, and 2) they are given lower weights otherwise. The global model can then be learned with these aggregated outputs to focus more on `confident' clients among those who have been trained on non-IID data. Technically, the similarity between client and distillation data can be obtained by learning an additional classifier that distinguishes between the two. In this way, we can derive the above confidence scores even from unlabeled distillation data.

To evaluate our approach, we conducted extensive experiments with multiple public datasets (MNIST, \mbox{CIFAR-10}, and \mbox{CINIC-10}~\cite{Darlow2018}) with various conditions. The results showed that the proposed DLAD provided promising performance gains compared with baselines in almost all the non-IID cases.

\section{Related Work}
\subsection{Learning from Decentralized Data}
Our main motivation is to learn a high-capacity machine learning model by leveraging decentralized data. FL~\cite{Mcmahan2017,Bonawitz2019} was proposed to address this problem by asking each client to train a shared model with their own data, while the server aggregates the client models to obtain a better global one. More recent work along this line of research extends the FL frameworks to be more communication efficient~\cite{Jeong2018,Konecny2016, Lin2017}, secure~\cite{Bagdasaryan2018,Bonawitz2017}, and applicable to a practical wireless setting~\cite{Giannakis2016,Wang2018,nishio2019client}. Another relevant work is Private Aggregation of Teacher Ensembles (PATE)~\cite{papernot2016semi,papernot2018scalable}, where multiple teacher (\ie, client) models trained with distinct data are aggregated and distilled to obtain a student (\ie, global) model, while preserving the data privacy by means of differential privacy~\cite{dwork2014algorithmic}. While PATE is potentially applicable to client models with heterogeneous architectures, the lack of adaptive aggregation of the client models in the distillation process will limit its performance when data are not IID, as will be demonstrated in our experiments.

\subsection{Distillation}
Network distillation~\cite{hinton2015distilling} was first proposed to compress and transfer recognition capabilities from one network to another. This idea was then extended to a variety of scenarios including but not limited to online distillation with multiple models~\cite{anil2018large}, semi-supervised learning~\cite{radosavovic2018data,tarvainen2017mean}, and reinforcement learning~\cite{rusu2015policy}. Some recent work has attempted to leverage distillation techniques for an FL setting~\cite{ahn2019wireless} and a collaborative learning setting~\cite{li2019fedmd}. However, to the best of our knowledge, only a very few studies have mentioned the problem of data non-iidness~\cite{li2019fedmd}, and they just averaged client models equally, which is insufficient to truly resolve the problem. Also, \cite{ahn2019wireless} requires distillation data to be fully annotated, whereas our work can accept non-labeled data.

\section{Preliminaries}
\subsection{Problem Setting}
Let $x\in \mathcal{X}$ be an input sample (\eg, images) and $y\in\mathcal{Y}$ be a ground-truth label annotated to input samples. In this work, we will focus particularly on classification problems; $\mathcal{Y}$ is given by a finite set of $L$ class labels: $\mathcal{Y}=\{1, \dots, L\}$.

To formulate the problem of decentralized learning, we consider the existence of a server and multiple clients. Suppose that $N$ clients $U_1,\dots,U_N$ each have their own labeled dataset $D_i=(X_i, Y_i),\; i \in \{1,\dots,N\}$, where $X_i=\{x^{(i)}_j\}$ and $Y_i=\{y^{(i)}_j\}$ and $y^{(i)}_j$ is a ground-truth class annotated to $x^{(i)}_j$. $D_i$ will be visible only to $U_i$, and will not be shared with the server nor the other clients $\{U_k \mid k\neq i\}$. Similar to the non-IID data condition evaluated in \cite{Mcmahan2017}, each client is supposed to observe a limited and dissimilar set of classes. This means that, for a set of observed classes $\mathcal{Y}^{(i)}$ where $y^{(i)}_j\in \mathcal{Y}^{(i)}\subset \mathcal{Y}$, $\mathcal{Y}^{(p)}$ and $\mathcal{Y}^{(q)}$, $p\neq q \in \{1, \dots, N\}$, \emph{are not necessarily the same}. Finally, we assume that each client has its own model (\emph{client model}) $M_i:\mathcal{X}\rightarrow\mathcal{Y}^{(i)}$ that was acquired using $D_i$, and this model will \emph{not} be updated during the training of a global model shown below.

Given a set of client models $\mathcal{M}=\{M_i\}$, our goal is to acquire a global model $M:\mathcal{X}\rightarrow\mathcal{Y}$ at the server side, \emph{which can classify samples of all the classes}.

\subsection{Network Distillation}
To transfer classification abilities from client models to the global one, we leverage the idea of network distillation~\cite{hinton2015distilling}. Suppose that we have one pre-trained model $M\sub{src}$ and another model $M\sub{tgt}$ that we will learn from scratch to inherit $M\sub{src}$'s classification ability. With another labeled dataset $D\sub{dist} = (X\sub{dist}, Y\sub{dist})$, we train $M\sub{tgt}$ so that it can imitate the outputs from $M\sub{src}$. This can be done by minimizing the following objective:
\begin{eqnarray}
    \mathcal{L}(D\sub{dist}) ={E}_{x\in X\sub{dist}}\left[l_1(M\sub{src}(x), M\sub{tgt}(x))\right]\nonumber \\
    + {E}_{x, y\in X\sub{dist}\times Y\sub{dist}}\left[l_2(M\sub{tgt}(x), y)\right], 
    \label{eq:distillation}
\end{eqnarray}
where $l_1, l_2$ are certain loss functions such as mean squared error and categorical cross entropy.

\section{Decentralized Learning via Adaptive Distillation}
\label{sec:dlad}
We extend the distillation objective in Eq.~(\ref{eq:distillation}) to make it applicable to our decentralized learning problem. Specifically, multiple client models $\mathcal{M}=\{M_i\}$ are provided as a distillation source, which is each trained with non-identical data $D_i=(X_i, Y_i)$. Moreover, we consider distillation data to be unlabeled $D\sub{dist}=X\sub{dist}$ to overcome the lack of public annotated data. These requirements, however, make it hard to apply existing approaches~\cite{anil2018large,ahn2019wireless,li2019fedmd} to our problem.

To this end, our proposed approach, DLAD, \emph{adaptively} aggregates outputs from the client models and uses the adaptive aggregation results to train a global model. Namely, our new objective is given as follows:
\begin{equation}
\small
    \mathcal{L}(X\sub{dist}) ={E}_{x\in X\sub{dist}}\left[l_1
\left(\sum_i w_i(x)M_i(x), M\sub{tgt}(x)\right)\right],
\label{eq:DLAD}
\end{equation}
where $l_1$ is categorical cross-entropy and $w_i(x)$ is a weight that satisfies $\sum_i w_i(x) = 1$. When each client model is trained from non-IID data and provides a variety of responses to $x\in X\sub{dist}$, $w_i(x)$ should be higher for the $i$-th client model that has observed similar samples of the same classes in their training data, and that can therefore be more confident as a teacher to inform the output to sample $x$. However, in our problem setting, it is not available which classes each $x$ in the distillation dataset belongs to, as well as which sets of classes each client data involves.

To compute $w_i(x)$ without knowing its labels, we first ask each client to train another binary classifier $C_i(x)\in [0, 1]$ with sigmoid outputs, to distinguish $X_i$ from $X\sub{dist}$. Similar to the discriminators trained in generative adversarial networks~\cite{goodfellow2014generative}, if this classifier is trained optimally, its output to sample $x$ is described as follows:
\begin{equation}
C^*_i(x) = \frac{p_i(x)}{p\sub{dist}(x) + p_i(x)},
\label{eq:classifier}
\end{equation}
where $p\sub{dist}(x)$ and $p\sub{i}(x)$ are the probability of sample $x$ in $X\sub{dist}$ and $X_i$, respectively. $C^*_i(x)$ can be represented by $C^*_i(x)=1-\frac{p\sub{dist}(x)}{p_i(x)+p\sub{dist}(x)}$. By fixing $x$ and regarding $p\sub{dist}(x)$ as a positive constant, $C^*_i(x)$ monotonically increases with $p_i(x)$ within $p_i(x)\in[0, 1]$. This means that $C^*_i(x)$ gives higher values when $x$ is more likely to be contained in $X_i$, and $M_i$ is confident about its output $M_i(x)$, accordingly.  To obtain $w_i(x)$, we compute the softmax on those classifier outputs, \ie,
\begin{equation}
    w_i(x)=\frac{\exp{C_i(x/T)}}{\sum_j\exp{C_j(x/T)}},
\label{eq:weight}
\end{equation}
where $T$ is a hyperparameter of temperature to control the smoothness of the output.

\section{Experiments}
We evaluated the DLAD on decentralized versions of multiple public image datasets. Since our goal was to evaluate how well our adaptive distillation algorithm worked on learning from non-IID data, we implemented all the training procedures in a single workstation for the simulation.

\subsection{Datasets}
As base datasets, we utilized MNIST, \mbox{CIFAR-10}, and \mbox{CINIC-10}~\cite{Darlow2018}. Note that \mbox{CINIC-10} is a challenging dataset because it comprises a large number of samples (270,000 in total) drawn from \mbox{CIFAR-10} and ImageNet. We decentralized them so that each of $N$ clients owns its subset with a limited number of classes, and a shared unlabeled dataset $X\sub{dist}$ is used for distilling client models into a global model. 

More specifically, for MNIST and \mbox{CIFAR-10}, we randomly chose 80\% of the samples (48,000 for MNIST and 40,000 for \mbox{CIFAR-10}) from the training dataset for $X\sub{dist}$. The remaining 20\% (12,000 for MNIST and 10,000 for \mbox{CIFAR-10}) became a client data pool. We chose this 80\%-20\% ratio to simulate real world conditions, where there is usually much more unlabeled data than labeled data. From the data pool, each client $U_i$ was supposed to randomly sample data of certain classes based on its predefined class probability $p_i$ to create its own training dataset which is $D_i=(X_i, Y_i)$. We want to have a sufficient number of training samples for each client, so the size of $D_i$ is set to half the size of the data pool (6,000 for MNIST and 5,000 for \mbox{CIFAR-10}, allowing duplicates). For \mbox{CINIC-10}, since it naturally includes a training dataset and a validation dataset (90,000 samples each), we assigned the whole validation set for $X\sub{dist}$, and the whole training set for the client data pool. Each client randomly sampled 20,000 samples from the data pool to create $D_i$.

As shown in Table~\ref{tab:class}, we tested four different types of data distribution to determine the robustness of the proposed method. Each type has different client-wise class probabilities (among ten classes). 
\begin{itemize}
\item \textbf{IID.} All clients follow the class probability of $p_i=[0.1, 0.1, \dots, 0.1]$, meaning that they have all the classes equally.
\item \textbf{Non-IID \#1 (NIID1).} Each client holds two consecutive classes. In this case $p_{5n+1} =[0.5, 0.5, 0, \dots,0]$, $p_{5n+2} =[0, 0, 0.5, 0.5, 0, \dots,0]$, and so on.
\item \textbf{Non-IID \#2 (NIID2).} All the clients share five consecutive classes (0--4), while holding one unique class. In this case $p_{5n+1} =[\frac{1}{6},\frac{1}{6},\frac{1}{6},\frac{1}{6},\frac{1}{6},\frac{1}{6},0,0,0,0]$, 
$p_{5n+2} =[\frac{1}{6},\frac{1}{6},\frac{1}{6},\frac{1}{6},\frac{1}{6},0,\frac{1}{6},0,0,0]$, and so on.
\item \textbf{Non-IID \#3 (NIID3).} Each client holds four classes: $p_{5n+1} =[0.25, 0.25, 0.25, 0.25, 0, \dots,0]$, $p_{5n+2} =[0.25, 0, 0, 0, 0.25, 0.25,0.25, 0, \dots,0]$, and so on.
\end{itemize}

\begin{table}[t]
\caption{The classes accessible by clients under different data distribution types}
\centering
\scalebox{1}{
\begin{threeparttable}
\begin{tabular}{lccccc}
\toprule
 & $U_{5n+1}$ & $U_{5n+2}$ & $U_{5n+3}$ & $U_{5n+4}$ & $U_{5n+5}$ \\
\midrule
IID & 0--9 & 0--9 & 0--9 & 0--9 & 0--9  \\
Non-IID \#1 & 0,1 & 2,3 & 4,5 & 6,7 & 8,9  \\
Non-IID \#2 & 0--4,5 & 0--4,6 & 0--4,7 & 0--4,8 & 0--4,9  \\
Non-IID \#3 & 0,1,2,3 & 0,4,5,6 & 1,4,7,8 & 2,5,7,9 & 3,6,8,9  \\
\bottomrule
\end{tabular}
\end{threeparttable}
}
\label{tab:class}
\end{table}

\begin{figure*}
    \centering
    \subfigure[IID]{%
        \includegraphics[clip, width=1\columnwidth]{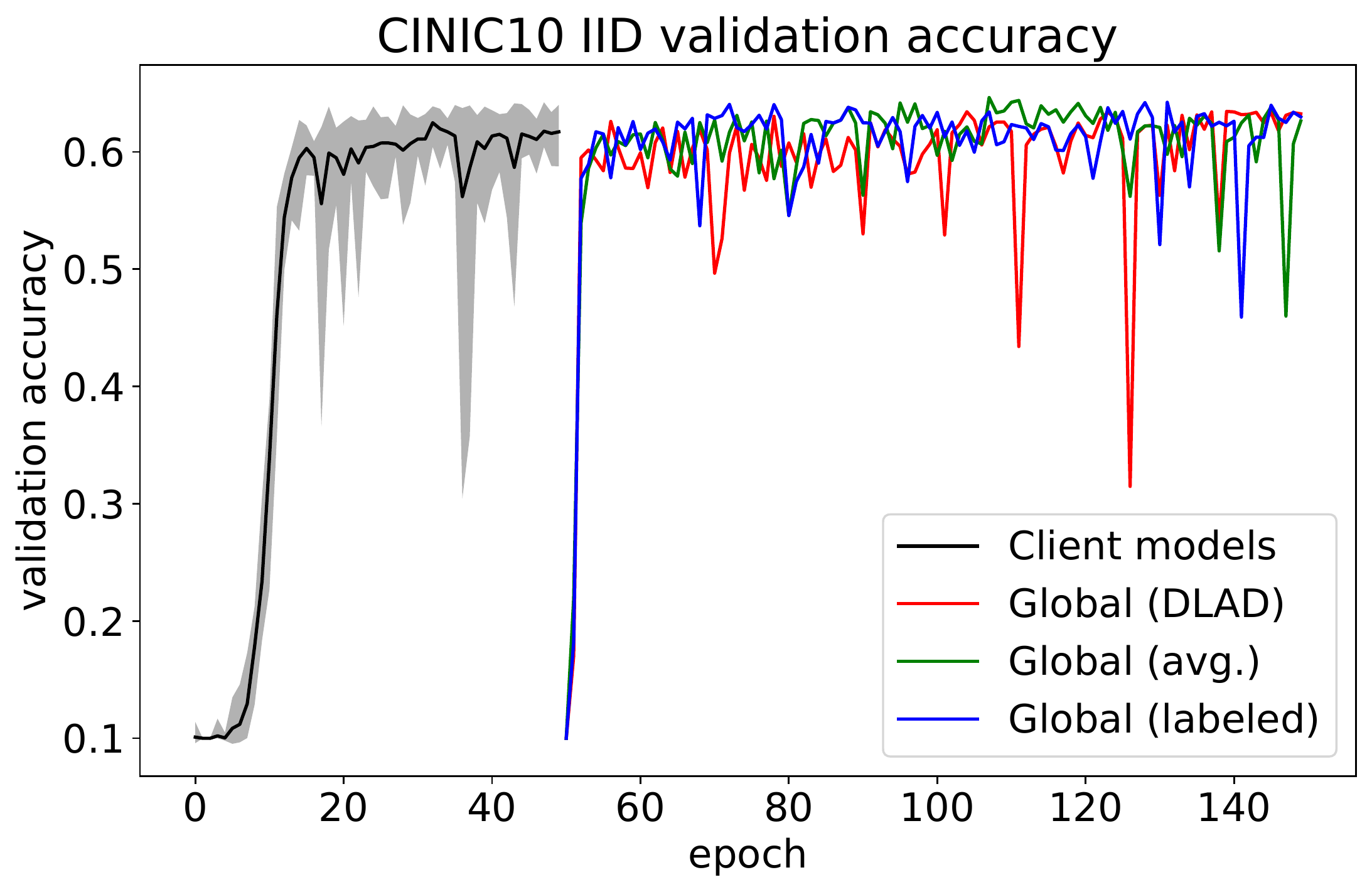}}%
        \quad
    \subfigure[NIID1]{%
        \includegraphics[clip, width=1\columnwidth]{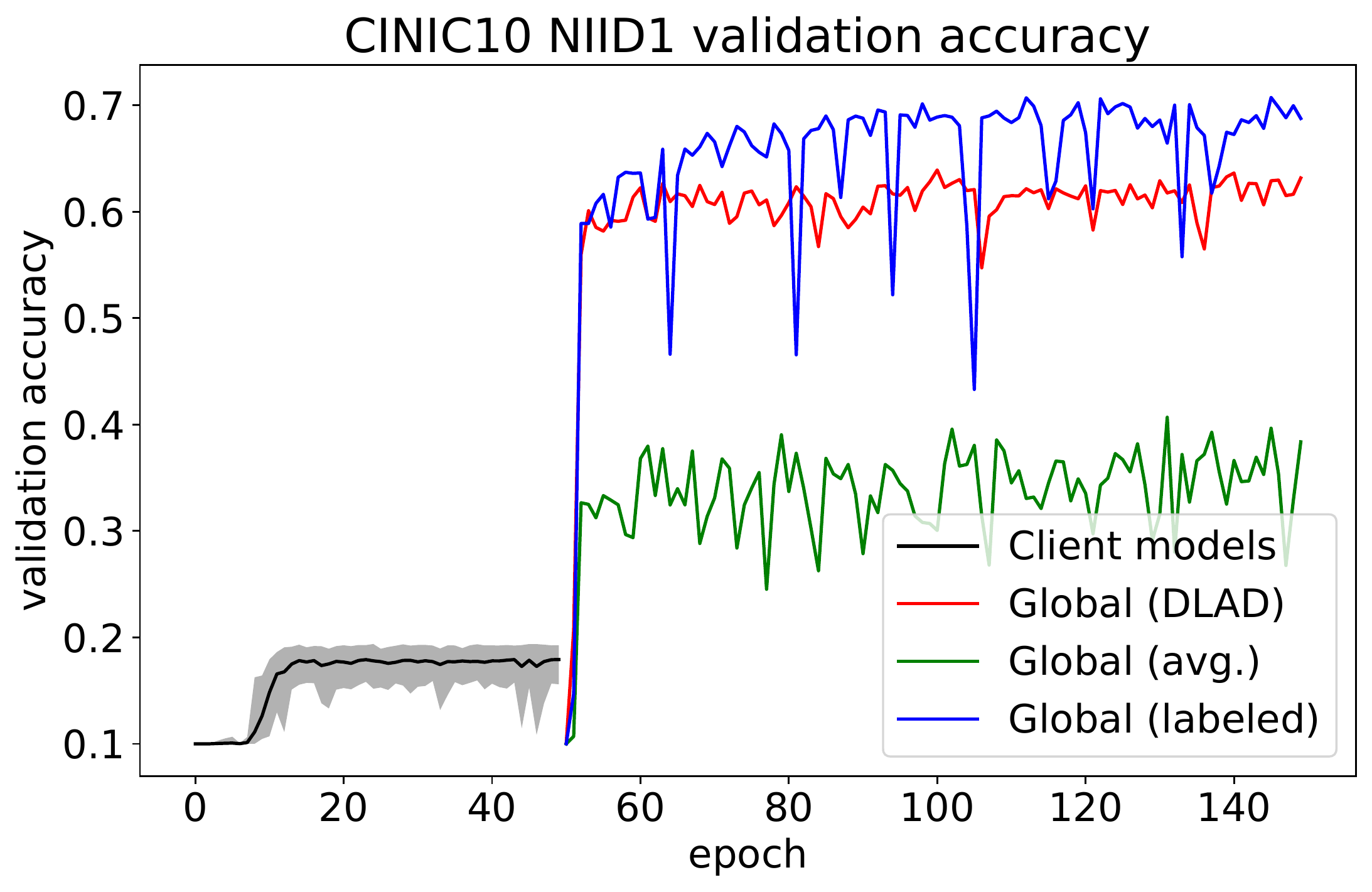}}%
        \quad
    \subfigure[NIID2]{%
        \includegraphics[clip, width=1\columnwidth]{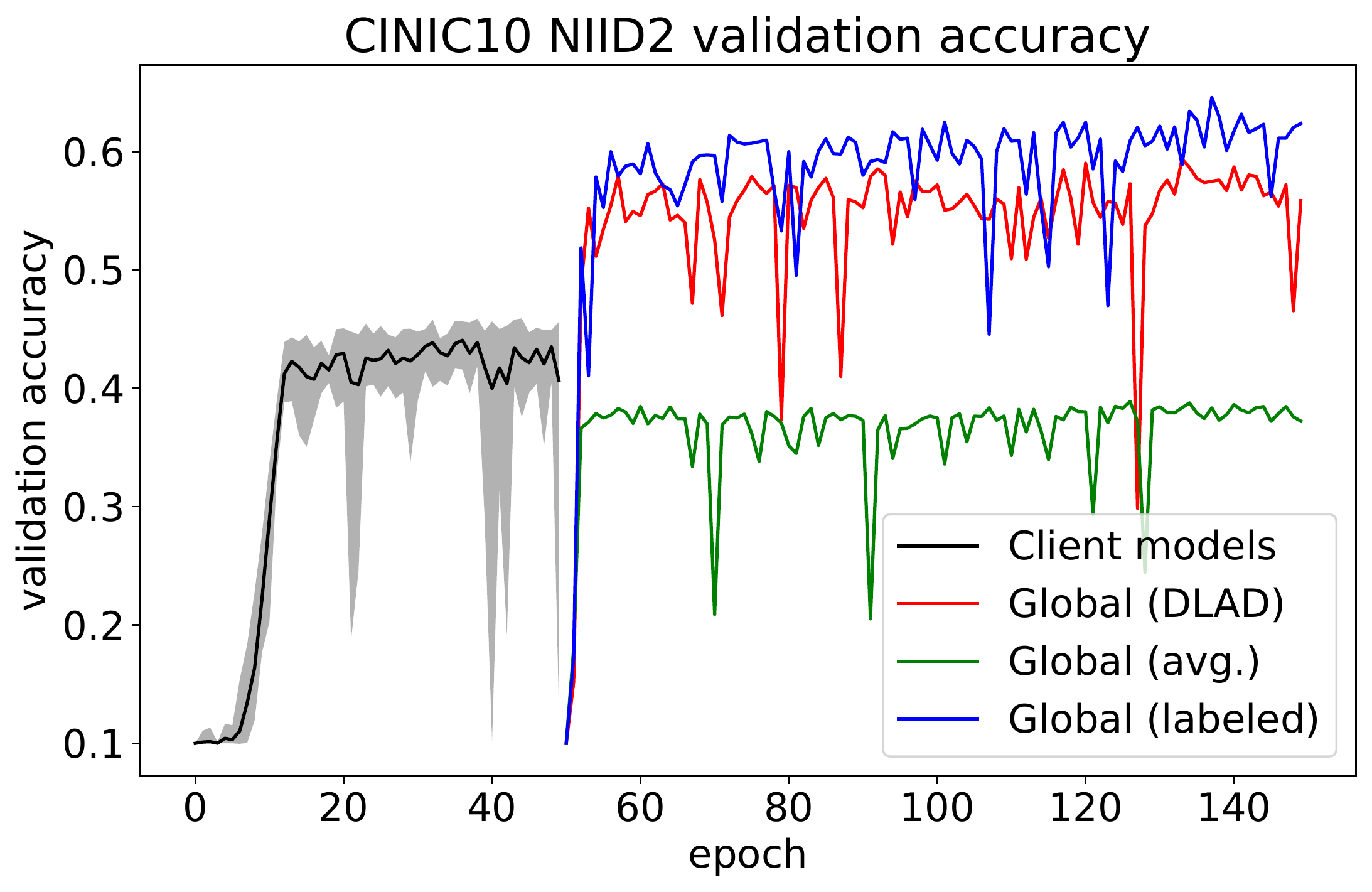}}%
        \quad
    \subfigure[NIID3]{%
        \includegraphics[clip, width=1\columnwidth]{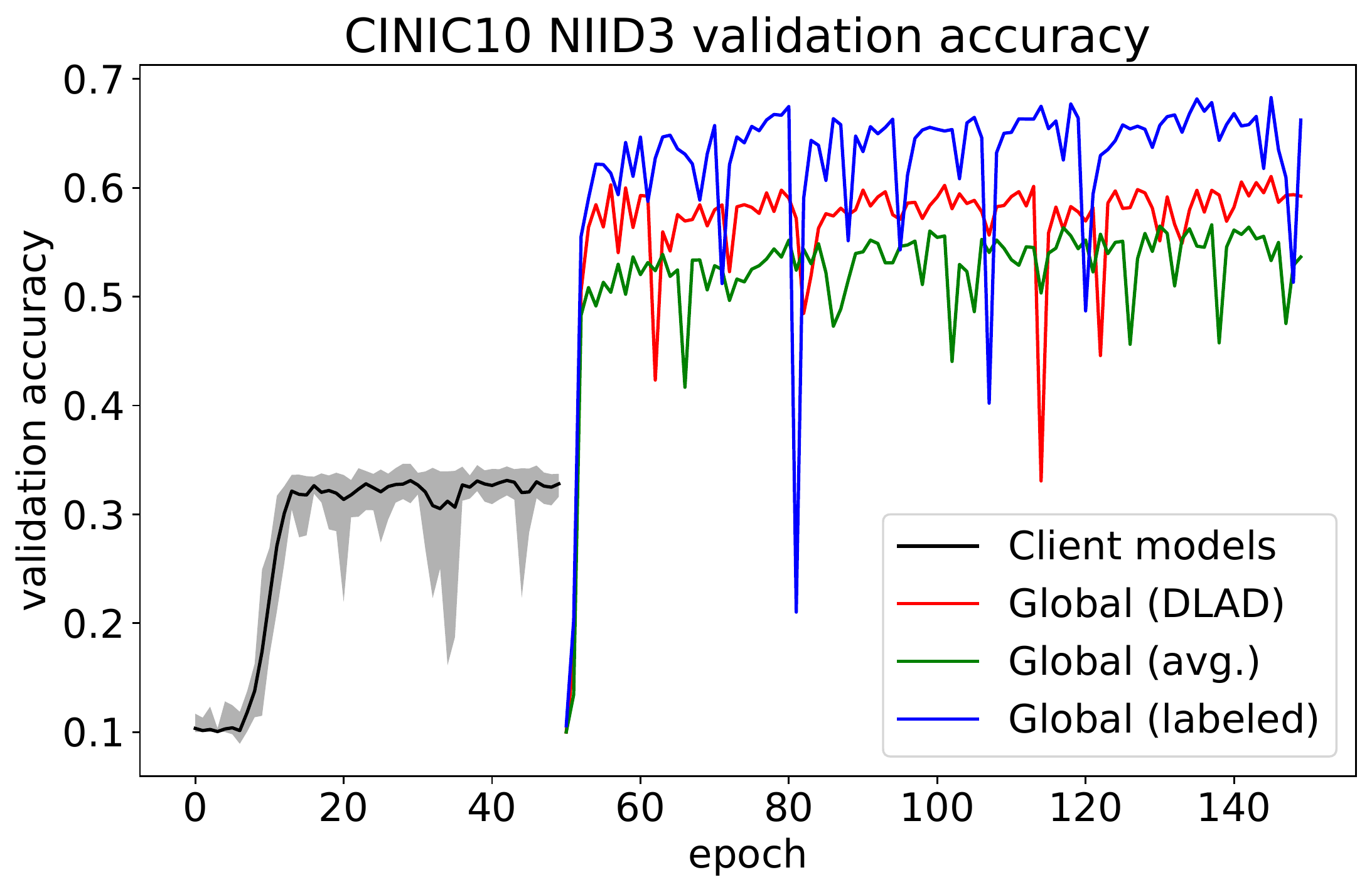}}%
    \caption{The results of Experiment 1 on CINIC-10. The first 50 epochs are the training process of clients (N=10). The last 100 epochs are the training (distillation) process of the global model. The proposed approach, Global (DLAD) highlighted in red, significantly outperformed the prior approach Global (avg.) shown in green and performed close to the upperbound (Global (labeled), shown in blue) even without labeled distillation data.}
    \label{fig:exp1}
\end{figure*}

\subsection{Implementation Details}

The whole training process consists of three steps: the training of client models, the training of binary classifiers, and the training of the global model.
\begin{itemize}
\item \textbf{Training client models.} Client models were either Deep Residual Network (ResNet18)~\cite{he2016deep} or Densely-connected Convolutional Networks (DenseNet)~\cite{huang2017densely}, where the former has a deeper architecture and thus is expected to perform better. Each model $M_i$ initially adopted weights pre-trained on ImageNet and then was trained with each client data $D_i$ using the Adam optimizer with the learning rate of 0.001 for 50 epochs with mini-batches of size 250.
\item \textbf{Training binary classifiers.} After the training of client models $M_i$ were finished, each binary classifier $C_i$ adopted the model architectures and the weights of $M_i$ and then was trained for 20 epochs. This step is necessary for estimating the aggregation weights (see Eq.~\ref{eq:classifier} and Eq.~\ref{eq:weight}). The optimization configuration was not changed, except that a sample weight of 1.5 was applied if the training sample was from $X_{i}$, which is to alleviate the effect of data imbalance ($X_{i}$ is much fewer than $X_{dist}$). 
\item \textbf{Training global model.} Finally, the global model $M$ also adopted weights of ImageNet and then was trained with the same optimization configurations for 100 epochs. A temperature $T$ of 0.05 was used for calculating the weight aggregation as in Eq.~\ref{eq:weight}.
\end{itemize}
During all the training steps, the input data $X_i$ and $X_{dist}$ were augmented by using the following parameters: rotation (20$^{\circ}$), shift in width, height, and color (0.2), and horizontal flip. All the implementations were done with Keras and evaluated on NVIDIA Tesla V100.

\subsection{Baselines and Metrics}
We compared the DLAD with a baseline that just averaged the outputs of multiple client models such as done in prior work~\cite{anil2018large,li2019fedmd}. As an evaluation metric, we computed the classification accuracy on the test subsets (predefined by each dataset). We also evaluated original performances of client models as well as the upper-bound performance of the global model when each $C_i$ performed optimally to distinguish $X_i$ from $X\sub{dist}$.

\begin{table*}[t]
\caption{Experiment 1: Comparing classification accuracy under different datasets and data distributions ($N=10$, ResNet for all models).}
\centering
\scalebox{1}{
\begin{tabular}{lcccccccccccccc}
\toprule
Dataset & \multicolumn{4}{c}{MNIST} & & \multicolumn{4}{c}{\mbox{CIFAR-10}} & & \multicolumn{4}{c}{\mbox{CINIC-10}}\\
Distribution & IID & NIID1 & NIID2 & NIID3 & & IID & NIID1 & NIID2 & NIID3 & & IID & NIID1 & NIID2 & NIID3\\
\cmidrule(lr){1-5} \cmidrule(lr){7-10} \cmidrule(lr){12-15}
Client (ResNet) & 0.9523 & 0.1990 & 0.6036 & 0.3947 & & 0.6354 & 0.1697 & 0.4267 & 0.3205 & & 0.6171 & 0.1792 & 0.4067 & 0.3280 \\
Client (DenseNet) & - & - & - & - & & - & - & - & - & & - & - & - & - \\
\cmidrule(lr){1-5} \cmidrule(lr){7-10} \cmidrule(lr){12-15}
Global (avg.) & 0.9806 & 0.3954 & 0.5164 & 0.9540 & & 0.7220 & 0.3648 & 0.4154 & 0.6099 & & 0.6256 & 0.3534 & 0.3804 & 0.5514 \\
Global (labeled) & 0.9836 & 0.9868 & 0.9845 & 0.9857 & & 0.7115 & 0.8127 & 0.7576 & 0.7755 & & 0.6256 & 0.6880 & 0.6183 & 0.6574 \\
\textbf{Global (DLAD)} & 0.9821 & 0.9820 & 0.9828 & 0.9840 & & 0.7314 & 0.6657 & 0.6847 & 0.7027 & & 0.6323 & 0.6266 & 0.5666 & 0.5934 \\
\bottomrule
\end{tabular}
}
\label{tab:exp1}
\end{table*}

\begin{table*}[t]
\caption{Experiment 2: Comparing classification accuracy under different model architectures ($N=10$, \mbox{CIFAR-10}, Non-IID \#1).}
\centering
\scalebox{1}{
\begin{tabular}{lcccccccc}
\toprule
Client model & \multicolumn{2}{c}{ResNet} & & \multicolumn{2}{c}{DenseNet} & & \multicolumn{2}{c}{ResNet/DenseNet}\\
Global model & ResNet & DenseNet & & ResNet & DenseNet & & ResNet & DenseNet\\
\cmidrule(lr){1-3} \cmidrule(lr){5-6} \cmidrule(lr){8-9}
Client (ResNet) & 0.1697 & 0.1697 & & - & - & & 0.1721 & 0.1721  \\
Client (DenseNet) & - & - & & 0.1837 & 0.1837 & & 0.1824 & 0.1824  \\
\cmidrule(lr){1-3} \cmidrule(lr){5-6} \cmidrule(lr){8-9}
Global (avg.) & 0.3021 & 0.2088 & & 0.4354 & 0.3786 & & 0.3648 & 0.2239  \\
Global (labeled) & 0.7991 & 0.8070 & & 0.8052 & 0.7816 & & 0.8127 & 0.7559  \\
\textbf{Global (DLAD)} & 0.5845 & 0.5683 & & 0.5856 & 0.5667 & & 0.6657 & 0.6642  \\
\bottomrule
\end{tabular}
}
\label{tab:exp2}
\end{table*}

\begin{table*}[!t]
\caption{Experiment 3: Comparing classification accuracy with increasing number of clients $N$ (\mbox{CIFAR-10}, ResNet for global model).}
\centering
\scalebox{1}{
\begin{tabular}{lccccccccccccc}
\toprule
Distribution & \multicolumn{13}{c}{Non-IID \#1} \\
Client model & \multicolumn{4}{c}{ResNet} & & \multicolumn{4}{c}{DenseNet} & & \multicolumn{3}{c}{ResNet/DenseNet}\\
No. of clients & 5 & 10 & 20 & 30 & & 5 & 10 & 20 & 30 & & 10 & 20 & 30 \\
\cmidrule(lr){1-5} \cmidrule(lr){7-10} \cmidrule(lr){12-14}
Client (ResNet) & 0.1721 & 0.1697 & 0.1696 & 0.1704 & & - & - & - & - & & 0.1721 & 0.1700 & 0.1704\\
Client (DenseNet) & - & - & - & - & & 0.1850 & 0.1837 & 0.1847 & 0.1839 & & 0.1824 & 0.1838 & 0.1818\\
\cmidrule(lr){1-5} \cmidrule(lr){7-10} \cmidrule(lr){12-14}
Global (avg.) & 0.3066 & 0.3102 & 0.3590 & 0.3857 & & 0.3471 & 0.4347 & 0.4719 & 0.4623 & & 0.3648 & 0.4907 & 0.4771\\
Global (labeled) & 0.7888 & 0.8000 & 0.8050 & 0.8160 & & 0.8197 & 0.8194 & 0.8255 & 0.8264 & & 0.8127 & 0.8252 & 0.8307\\
\textbf{Global (DLAD)} & 0.6376 & 0.5838 & 0.6227 & 0.6553 & & 0.4725 & 0.5784 & 0.6251 & 0.5748 & & 0.6657 & 0.6379 & 0.6470\\
\midrule
Distribution & \multicolumn{13}{c}{Non-IID \#2} \\
Client model & \multicolumn{4}{c}{ResNet} & & \multicolumn{4}{c}{DenseNet} & & \multicolumn{3}{c}{ResNet/DenseNet}\\
No. of clients & 5 & 10 & 20 & 30 & & 5 & 10 & 20 & 30 & & 10 & 20 & 30 \\
\cmidrule(lr){1-5} \cmidrule(lr){7-10} \cmidrule(lr){12-14}
Client (ResNet) & 0.4331 & 0.4267 & 0.4205 & 0.4194 & & - & - & - & - & & 0.4331 & 0.4236 & 0.4199\\
Client (DenseNet) & - & - & - & - & & 0.4694 & 0.4339 & 0.4471 & 0.4533 & & 0.3985 & 0.4330 & 0.4433\\
\cmidrule(lr){1-5} \cmidrule(lr){7-10} \cmidrule(lr){12-14}
Global (avg.) & 0.4163 & 0.4069 & 0.4185 & 0.4103 & & 0.4216 & 0.4175 & 0.4182 & 0.4270 & & 0.4176 & 0.4190 & 0.4199\\
Global (labeled) & 0.7383 & 0.7659 & 0.7728 & 0.7706 & & 0.7774 & 0.7506 & 0.7987 & 0.7957 & & 0.7625 & 0.7993 & 0.7632\\
\textbf{Global (DLAD)} & 0.6660 & 0.6772 & 0.6952 & 0.6914 & & 0.6170 & 0.6151 & 0.6786 & 0.6904 & & 0.6219 & 0.6782 & 0.7025\\
\midrule
Distribution & \multicolumn{13}{c}{Non-IID \#3} \\
Client model & \multicolumn{4}{c}{ResNet} & & \multicolumn{4}{c}{DenseNet} & & \multicolumn{3}{c}{ResNet/DenseNet}\\
No. of clients & 5 & 10 & 20 & 30 & & 5 & 10 & 20 & 30 & & 10 & 20 & 30 \\
\cmidrule(lr){1-5} \cmidrule(lr){7-10} \cmidrule(lr){12-14}
Client (ResNet) & 0.3156 & 0.3205 & 0.3167 & 0.3154 & & - & - & - & - & & 0.3156 & 0.3169 & 0.3135\\
Client (DenseNet) & - & - & - & - & & 0.3373 & 0.3063 & 0.3237 & 0.3242 & & 0.2752 & 0.3079 & 0.3100\\
\cmidrule(lr){1-5} \cmidrule(lr){7-10} \cmidrule(lr){12-14}
Global (avg.) & 0.5737 & 0.6260 & 0.5786 & 0.6315 & & 0.5740 & 0.5570 & 0.5685 & 0.5879 & & 0.5489 & 0.6172 & 0.6366\\
Global (labeled) & 0.7930 & 0.8015 & 0.8052 & 0.8159 & & 0.7872 & 0.7898 & 0.7923 & 0.8196 & & 0.8088 & 0.7981 & 0.8190\\
\textbf{Global (DLAD)} & 0.6842 & 0.6926 & 0.6978 & 0.7001 & & 0.6502 & 0.6197 & 0.6456 & 0.6560 & & 0.6055 & 0.6820 & 0.7060\\
\bottomrule
\end{tabular}
}
\label{tab:exp3}
\end{table*}

\subsection{Results}

We conducted a comprehensive set of experiments to examine the effects of different datasets, architectures, and numbers of clients. Tables~\ref{tab:exp1}, \ref{tab:exp2}, and \ref{tab:exp3} list the results, which are the classification performances of each model showing the median value of the test accuracy of the last ten epochs. Overall, we found that the global model obtained substantially higher performances than client models, thus demonstrating the effectiveness of involving multiple clients in the training. 
Moreover, in all the non-IID cases, the proposed DLAD significantly outperformed the baseline method (as shown in the ``Global (avg.)'' rows) that averaged client models without confidence weights, and sometimes performed comparably well with the upper-bound performances (as shown in the ``Global (labeled)'' rows). 
These results suggest the effectiveness of using our adaptive aggregation strategy to resolve non-IID problems. This is a novel and important contribution of our proposed DLAD method, as well as an advantage over other studies considering the fact that recent work on distillation and FL~\cite{ahn2019wireless,li2019fedmd} only incorporated a simple average aggregation procedure that is similar to our baseline method.

\subsubsection{Effect of Datasets}
In the first experiment, we investigated whether the proposed DLAD works well on different datasets and different data distributions. We fixed the model architecture to ResNet and the number of clients to 10, then experimented on three datasets (MNIST, \mbox{CIFAR-10}, \mbox{CINIC-10}) and four data distribution types (IID, NIID1, NIID2, NIID3). In Table~\ref{tab:exp1}, we can see that for the MNIST dataset, as each client only held two (NIID1), six (NIID2), or four classes (NIID3), the client model accuracy predictably converged at 0.2, 0.6, and 0.4, respectively. The proposed DLAD distilled knowledge from multiple client models, so it was not affected by non-iidness and could achieve accuracy higher than 0.98 in all MNIST experiments. For the other two datasets, \mbox{CIFAR-10} and \mbox{CINIC-10}, as the task difficulty increased, the final performance of DLAD decreased but was still significantly better than the baseline. 
Fig.~\ref{fig:exp1} shows the experimental results on \mbox{CINIC-10}, which is the most difficult task. From the figure we can observe that, for the IID case, since the client models have the same architecture and the same data distribution, distillation did not bring any change to the model performance. However, for the NIID cases, while the baseline distillation method yielded a moderate increase or even a decrease (see NIID2) in the performance, the proposed DLAD method yielded an overall good performance.

\subsubsection{Effect of Network Architectures}
Next, we investigated whether the DLAD works well on different model architectures. We fixed the dataset to \mbox{CIFAR-10}, the data distribution type to NIID1, and the number of clients to 10, then experimented on six combinations of different model architectures. Specifically, the client model architectures could be all ResNet, all DenseNet, or ResNet/DenseNet (50\%-50\%), while the global model architecture could be either ResNet or DenseNet.
An advantage of DLAD is that it applies no restriction on the model architectures of clients, which allows the clients to customize their own models to meet distinct specifications on their hardware and software.
As shown in Table~\ref{tab:exp2}, regarding the choice of client model architectures, the ResNet/DenseNet case outperformed the other two homogeneous cases, which indicates that diversity of client models may benefit the robustness of DLAD. For the global model, ResNet performed better than DenseNet, which can be ascribed to its deeper architecture.

\subsubsection{Effect of Number of Clients}
Finally, we explored the effect of increasing the number of clients, as well as further verifying the performance. We fixed the dataset to \mbox{CIFAR-10} and the global model to ResNet, then experimented on four cases of different numbers of clients ($N=5,10,20,30$), three cases of client model architectures (all ResNet, all DenseNet, and ResNet/DenseNet), and three cases of data distribution (NIID1--3). In Table~\ref{tab:exp3}, we can see that 1) the performance of DLAD was consistent under various experimental conditions and 2) with only a few exceptions, the performance of DLAD was generally improved by involving more clients. 
These results indicate that the proposed method is suitable for large-scale usage, though we may also need to address the increasing cost of server-client communications.

\subsubsection{Limitations}
Currently, our work has two limitations. First, although DLAD allows a global model to be trained with unlabeled data, $X\sub{dist}$, this distillation source needs to be drawn from relevant domains. If all the classes in $X\sub{dist}$ are different from those of $Y_i$, that may make all the outputs from $C_i$ lower and weighted aggregation unhelpful. Nevertheless, our approach can now remove the annotation cost for distillation from client models, which has been required in prior work on decentralized learning~\cite{ahn2019wireless,li2019fedmd}. Secondly, this work did not address some practical aspects of learning from decentralized data, such as communication costs and security concerns. Another interesting direction for future work is to incorporate secure aggregation into our framework, as done in FL~\cite{Bonawitz2017}.

\balance

\section{Conclusion}
We have presented DLAD, a new decentralized learning approach designed to leverage multiple client models of different architectures, which have been acquired using non-IID data. We have reported promising results with experiments on multiple public datasets, where the DLAD outperformed a conventional distillation-based approach that has widely been used in prior work such as \cite{ahn2019wireless,li2019fedmd}. Our work will be of interest in multiple domains where one wishes to obtain knowledge from multiple clients who own relevant but dissimilar models. Future work will seek to extend our approach to various tasks beyond visual recognition such as natural language processing~\cite{Hard2018} and medical applications~\cite{Wenqi2019}.

\bibliographystyle{IEEEtran}

\end{document}